# Chapter 14: Looking Forward: Challenges and Opportunities in Agentic AI Reliability

Liudong Xing[1], Janet (Jing) Lin[2]

[1]Electrical and Computer Engineering Department, University of Massachusetts, Dartmouth, USA

[2]Luleå University of Technology, Luleå, Sweden

**Abstract** This chapter presents perspectives for challenges and future development in building reliable AI systems, particularly, agentic AI systems. Several open research problems related to mitigating the risks of cascading failures are discussed. The chapter also sheds lights on research challenges and opportunities in aspects including dynamic environments, inconsistent task execution, unpredictable emergent behaviors, as well as resource-intensive reliability mechanisms. In addition, several research directions along the line of testing and evaluating reliability of agentic AI systems are also discussed.

## 1 Introduction

The AI conversation can be traced as far back as Alan Turing's milestone paper published in 1950, which considered the fundamental question "Can machines think?" [1]. In 1956, AI got its name and mission as a scientific field at the first AI conference held at Dartmouth College [2]. Following AI's foundational period in the 1950s ~ 1970s, AI has evolved from early rule-based systems (1970s ~ 1990s), through classical machine learning and deep learning with neural networks (1990s ~ 2020s), to today's generative and agentic AI systems (since 2010s). Correspondingly, as a vital requirement of these systems, the reliability concept and concerns are also evolving, particularly in the interpretation of "required function" (see Table 1 in Chapter 10), based on the definition in standards like ISO 8402 "*The ability of an item to perform a required function, under given environmental and operational conditions and for a stated period of time*". While a conventional AI system is concerned with providing stable and accurate classifications, predictions, or optimizations, a reliable generative AI system focuses on producing outputs that are trustworthy, consistent, safe, and contextually appropriate [3]. Building on both, a reliable agentic AI system should additionally conduct functions of reasoning, goal alignment, planning, safe adaption and interaction in dynamic and collaborative multi-agent contexts.

The expansion of reliability concepts has introduced new challenges and research opportunities, as exemplified in Fig. 1. In the following sections, we shed lights on these challenges and opportunities in building reliable AI systems, particularly, agentic AI systems.

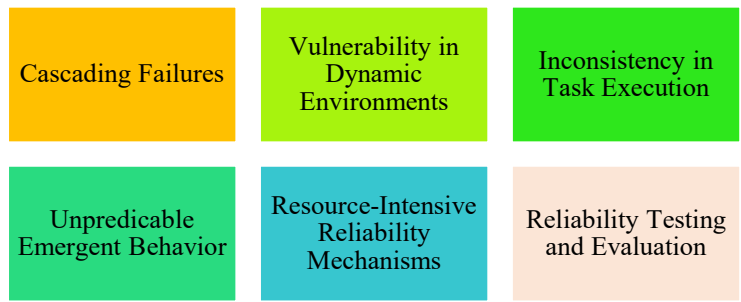

**Fig. 1** Research Challenges and Opportunities in AI Reliability

## 2 Cascading Failures

Cascading failures abound in diverse application domains and have been extensively studied for power grids [4-6] and various Internet of Things (IoT) systems [7-9]. Physical or logical interdependence typically exists among components in these systems, where the malfunction of one component may trigger a cascade of unexpected and often undesired behavior or state changes of other components, prompting domino chain effects. Without effective and timely control and mitigation mechanisms, cascading failures can cause catastrophic consequences [10]. As AI systems grow more complex, they become increasingly vulnerable to cascading failures due to intra-layer and inter-layer interacting components, opaque decision-makings, as well as insufficient cross-layer visibility where assumptions, errors, state changes, or unexpected behavior in one layer may not be adequately detected and mitigated or communicated with other layers.

Specifically, in Chapter 10, an eleven-layer failure stack is presented to organize potential vulnerabilities from the physical and computational foundational layers (hardware, power, system software, AI frameworks), to core intelligence layers (models, data), to operational layers (applications, execution, monitoring, learning), and to the top agentic layer. Cascading failures may happen at the same layer where, for example, errors in one agent can cascade across dependent agents incurring system-wide disruptions. They can also occur vertically, propagating upward or downward across different layers. For example, poor agentic choices can exacerbate or amplify lower-layer vulnerabilities, spreading failures downward. Since upper-layer functions are inherently contingent on the integrity and reliability of lower layers, upward cascades may occur more likely. For example, errors or bugs in system software can render the AI framework (libraries and pipelines) unusable; hardware malfunctions, unstable learning, and biased dataset can destabilize autonomy and compromise decision making at the agentic layer.

Correlated vulnerabilities driving cascading failures highlight that AI system reliability cannot be ensured in isolation at individual layers. Single-layer robustness is no longer sufficient; instead, a cross-layer framework with systematic coordination is essential. The development of such a framework involves several key research problems, as illustrated in Fig. 2 and explained below, each of which presents significant opportunities for advancing reliability and resilience of agentic AI systems.

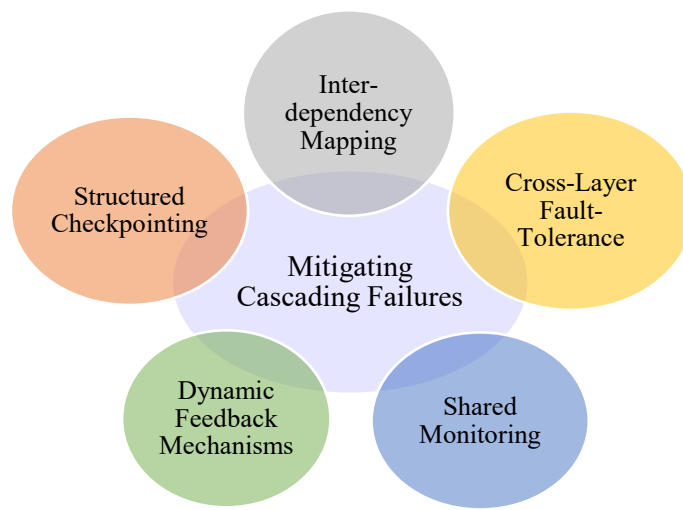

**Fig. 2** Key Open Research Problems in Mitigating Cascading Failures

**Cross-Layer Interdependency Mapping:** To enable systematic coordination, it is pivotal to explicitly map dependencies across different layers in the stack through, for example, tracing how upper layers are functionally dependent on lower layers and identifying potential propagation paths across the hierarchy. The goal of such mapping is to enable targeted and strategic interventions and facilitate early detection of vulnerable points.

**Dynamic Feedback Mechanisms:** Agentic AI systems need to learn and adapt in real-time and thus must incorporate dynamic feedback loop mechanisms to help the system stay resilient in changing or adversarial environments. The mechanisms should flow both upward and downward across the hierarchy. Specifically, a higher layer should be able to flag concerns about the performance of an AI model that may prompt data correction or re-training. A lower layer should adapt its behavior in response to evolving objectives or corrective feedback signals provided by higher layers.

**Cross-Layer Shared Monitoring:** Unlike traditional monitoring that primarily focuses on observing performance of individual components or layers, the monitoring in the cross-layer approach should be able to detect correlated anomalies spanning multiple layers that would otherwise remain hidden in the case of each layer being monitored separately. Additionally, it is vital to collect telemetry that can capture behavioral dependencies across layers, for example, how data drift impacts model confidence and subsequently affects decision-making patterns. To surface correlated anomalies and reveal cross-layer interactions, shared observability platforms must be developed.

**Cross-Layer Fault-Tolerance:** Redundancy (hardware, software, information and time) has long been used in traditional systems to contain and tolerate failures. Likewise, AI systems should implement both intra-layer and cross-layer redundancy to isolate faults and limit the impact of cascading failures. For example, redundant reasoning might be implemented at the agentic layer to cross-validate conclusions using different independent mechanisms, reducing reliance on any single, potentially unreliable data source.

**Structured Checkpointing:** To mitigate the risk of flawed or erroneous inputs spreading upward unchecked, well-structured checkpoints could be implemented to assess the health status at critical cross-layer junctures, which can trigger layer-specific or system-wide interventions. For example, an agentic planner should not trust the output of an AI model unless it passes certain robustness checks. These checks may include, for instance, using uncertainty thresholds to assess the confidence level of model predictions. In the case of the model exhibiting high uncertainty, the output may be flagged for human engagement, or the fallback mechanism can be triggered. Such intervention may prevent unreliable inferences from impacting higher-layer actions or decisions. Establishing checkpoints coupled with robustness checks form a defensive barrier that can prevent untrustworthy outputs at one layer from spreading to other layers, especially upward into decision-making layers with high-stake or irreversible consequences.

In addition to technical challenges and research opportunities explained above, effective coordination among teams working on different layers is essential to align on design assumptions, potential failure modes, and risk models. System architecture should facilitate such integration, avoiding rigid modular boundaries that conceal cross-layer dependencies and cascading failure risks.

## 3 Vulnerability in Dynamic Environments

Characterized by autonomy, goal-directed behavior, and the capability of acting in open-ended contexts, agentic AI systems can be unpacked in terms of several interrelated challenges arising from the dynamic environments [11, 12], as summarized in Fig. 3.

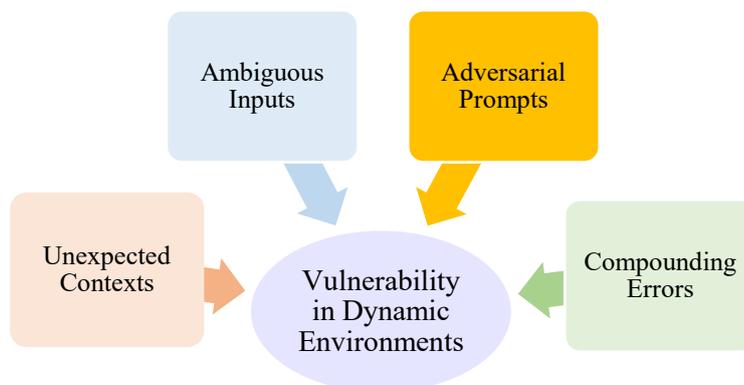

**Fig. 3** Challenges in Dynamic Environments

**Adapting to Unexpected or Unpredictable Contexts:** Due to novel objects introduced, revised rules or policies, or emergent behaviors, an agentic AI system can misinterpret its surroundings, fail to generalize rules or learned behaviors to new conditions, or make unsafe decisions in the event of edge cases. The unexpected changes in real-world conditions require re-adaptation that the AI system was not explicitly trained for.

**Ambiguous Inputs:** Instructions or prompts given by human users are often incomplete, ambiguous or even contradictory. The interpretation of such prompts requires deep context awareness and often relies heavily on human intent and norms, which AI systems do not inherently possess. As a result, AI systems may interpret ambiguous prompts literally, leading to actions that are technically correct but not aligned with human intention. Even worse, the system may fill in missing information by generating incorrect content, which can manifest as hallucinations or inappropriate behaviors.

**Adversarial Prompts:** In addition to ambiguous and underspecified prompts, agentic AI systems can be susceptible to malicious prompts or inputs, which are crafted to elicit unintended behavior or bypass safeguards. Even in non-malicious contexts, misaligned or harmful behaviors may be triggered based on how the AI interprets and extrapolates from its input.

**Compounding Errors:** Because agentic AI systems are often tasked with long-horizon, multi-step decision-making in dynamic environments, they are prone to compounding errors. Specifically, a misinterpretation or a small mistake occurring in an early step can cascade or accumulate, leading to compounding failures even without user awareness.

To address the above challenges of agentic AI systems in the face of novelty, ambiguity, and maliciousness, potential research opportunities in causal modeling and reasoning, intent modeling, explainability and transparency, and robustness to shifts and adversarial inputs could be explored, contributing to more robust, transparent, and adaptable decision-making in dynamic environments.

More specifically, it should be pivotal to equip an agentic AI system with the capability of reasoning about cause-effect relationships, rather than just pattern recognition and matching. With such capability, agents could predict the outcomes of their actions and adapt policies to changing environments to prevent or mitigate unintended consequences. To achieve these, future research may explore causal discovery [13, 14] in unstructured and high-dimensional data like language and perception and integrate causal models into the decision-making pipeline. Moreover, it is important to enable AI agents to reason about human preferences and beliefs even when they are under-specified; pragmatic reasoning and social reasoning could be explored to infer human intent, avoiding misaligned or harmful behaviors. Another research path could involve making the reasoning and decision processes of AI agents transparent and understandable to human users through, for example, explicit chain-of-thought traces in multi-step reasoning, dynamic dialogue between users and agents, and introspection where an agents can explain its plans and goals.

To ensure that an agentic AI system remains reliable under changing or new conditions, research questions such as out-of-distribution detection and incorporating epistemic uncertainty into action selection and planning must be addressed. To build system's resilience to adversarial inputs, it should be promising to explore automated red-teaming to discover edge-case vulnerabilities and develop context-sensitive sandboxing strategies for managing high-risk AI behaviors.

## 4 Inconsistent Task Execution

Agentic AI systems are expected to plan and coordinate subtasks (possibly across multiple agents) to produce goal-aligned behavior reliably. However, in practice, they often display variability in their outputs [15-17]. Specifically, outputs vary across repeated runs for the same task or among agents handling similar subtasks, undermining reliability, reproducibility, predictability and agent coordination. The inconsistency in task execution can be attributed to factors such as stochastic generation, underspecified task prompts, non-deterministic tool use (search engines, APIs), drift in model behavior due to adaptive learning or fine-tuning, and unclear policies for executing repeated or overlapping tasks.

To ensure consistent task execution in agentic AI systems, a promising research direction is the development of integrated methods that encompass grounded reasoning, explicit action-validation cycles, and stabilized memory (Fig. 4). Such a combined solution method could lead to more reliable, reproducible, and trustworthy AI agents, promoting consistency in agent behaviors, actions and decisions. Its three interrelated components are elaborated below.

**Grounded Reasoning:** Through integrating external knowledge retrieval (documents, policies, prior actions or decisions) into the reasoning pipeline of AI agents, outputs can be grounded in static and verifiable sources rather than relying solely on stochastic generation or internal memory. More specifically, the grounded reasoning can be achieved by retrieving relevant context from indexed knowledge documents or sources, injecting the retrieved context directly into agent prompts or planning modules, and integrating it with memory recall to ensure coherence across task executions.

**Action-Validation Cycle**s: Instead of generating a one-shot response, AI agents conduct reasoning and actions in iterative and reflective loops, where checkpoints are introduced to allow an agent to review its state and output, or make alignment with past decisions or outputs that can be retrieved at each step. In the case of inconsistencies being detected, the step may be retried. Such self-validation and dynamic plan adjustments could lead to improved state consistencies across iterations.

**Stabilized and Shared Memory**: To enable ground reasoning and action-validation loops, a stabilized memory mechanism is essential, which supports the retrieval of past assumptions, plans or decisions within each loop and the examination of causal relationship between past and current reasoning steps. The mechanism should also facilitate coordination among multiple agents via context sharing; the agents should be able to access common retrieval indexes and shared memory, ensuring they execute tasks from the same base of knowledge and thereby enabling task reproducibility and cross-agent consistency.

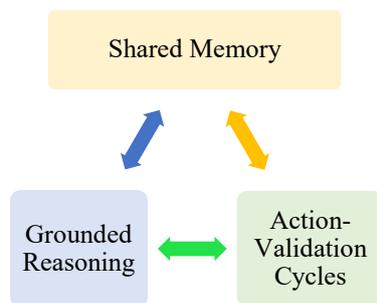

**Fig. 4** Integrated Methodology for Addressing Inconsistency

## 5 Unpredictable Emergent Behavior

In agentic AI systems, multiple autonomous agents often need to interact in complex and open-ended environments, potentially leading to unexpected or misaligned system-level outcomes, even when each individual agent behaves as intended [18, 19]. More specifically, different agents are autonomous, pursuing their goals independently in shared environments. However, they are also interactive and adaptive during task executions, leading to changes to the shared environments. Such interaction effects can create behaviors that are not intended or foreseen by any single agent and may not emerge until real-time executions. These behaviors are not explicitly programmed, are often resistant to traditional alignment or debugging tools, cannot be easily predicted based on local rules, and are often difficult to reverse once in motion. Consequently, they can be harmful and misaligned with human intent or values. Examples include destructive competition over shared resources, system-level goal mis-generalization, feedback loops amplifying small effects into catastrophic failures, unintended cooperation or collusion undermining fairness and trust, and cross-agent misalignment causing goal divergence and non-traceable harmful behavior.

Addressing emergent behaviors needs to overcome several difficulties in non-linearity, lack of global perspective, and scale effects. To elaborate, the global behavior of the system cannot be fully understood by examining its individual agents in isolation as it is more than the sum of its parts. Small, insignificant changes in one module can cascade, leading to nonlinear and often unpredictable shifts in overall system behavior. Also, each agent acts based on its local goals, observations, and rules; no single agent has a global view of the entire system, making cross-agent coordination fragile. Additionally, some emergent behaviors only manifest at scale during real-time simulation or deployment, making them difficult to be detected during the development stage.

To mitigate the risk of unpredictable emergent behaviors, there is a need to treat a multi-agent AI system as a complex ecosystem with system-level monitoring and causal tracing, shared ontologies for inter-agent communications, coordination protocols for preventing conflicting behaviors, and collective incentive design for ensuring system-level alignment. Along this research direction, it is also significant to implement redundancy, fallback strategies, and self-healing mechanisms to manage unexpected emergent risks. In particular, the development of such an ecosystem involves addressing key open research problems, as outlined in Fig. 5 and detailed below.

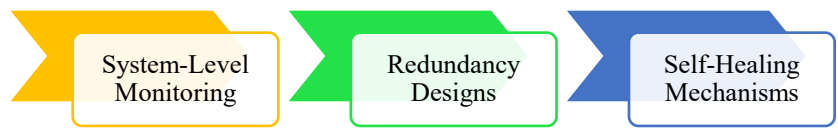

**Fig. 5** Key Research Opportunities for Addressing Emergent Behavior

**System-Level Monitoring:** The goal is to design real-time monitoring systems that can observe and evaluate agent behaviors both individually and collectively, and trigger interrupts to contain damage when anomalies are detected. Possible research problems include detecting emergent patterns or behavioral deviation from expected norms; summarizing multi-agent states through aggregating and interpret metrics reflecting system-wide behaviors like resource usage, quality of coordination, and divergence from expected plans; designing safety-critical interrupt signals that can override local agent autonomy, automatically halt or adjust agent behaviors, or trigger system rollback or fallback strategies in the event of anomalies or failures being detected.

**Redundancy Designs:** The goal is to ensure that the system can continue to function even in the presence of agent or planning module failures. Potential research problems include designing passive and active redundancy techniques [20] that can restore system functionality in failure cases; and developing graceful degradation techniques that enable agents to continue their functionality but with degraded performance through, for example, shifting to more conservative or simpler policies when optimal performance cannot be maintained under unexpected stresses.

**Self-Healing Mechanisms:** The goal is to design mechanisms for post-failure correction or repair. Related research problems include designing self-correction mechanisms that can reverse recent actions causing the harmful or error states; developing rollback strategies that can revert to a previous safe state or decision path; and investigating distributed recovery agents capable of watching and repairing coordination failures.

## 6 Resource-Intensive Reliability Mechanisms

To ensure reliability and mitigate risks such as cascading failures and emergent behaviors discussed in prior sections, reliability mechanisms are essential components of AI systems where redundancies in the form of hardware, software, information, and time [20] are built in to prevent failures or harmful outcomes. However, these mechanisms are resource-intensive, particularly involving high computational overhead, energy consumption and infrastructure cost. Extra computational overhead arises from simultaneously running multiple redundant processes or models that consumes more GPU/CPU cycles and performing frequent consistency checks and validation steps. The excessive energy consumption results from the operation of additional hardware resources as well as continuous monitoring and fail-safe mechanisms, which drives up operational costs. Moreover, the extra resources (servers, data centers) increase the demand for cooling and maintenance, thereby raising overall infrastructure costs as well. All these factors can make AI systems expensive and less accessible, especially in resource-limited settings. Fig. 6 outlines several potential mitigation strategies to address the resource-intensive reliability challenge in agentic AI systems, which are explained below.

**Adaptive Redundancy:** Instead of maintaining uniform redundancy and reliability-related checks, the redundancy mechanism can be dynamically scaled and adjusted based on real-time risk levels and energy availability. In the case of low risk (e.g., low-stakes task, stable environment) or limited energy, the level of redundancy or frequency of checks can be reduced to save energy consumption and operational cost. In the case of high risk (e.g., uncertain conditions, critical and complex operations) or energy-rich environment, reliability measures can be ramped up.

**Model Pruning:** By maintaining robustness in critical components, redundant or non-essential parts of the AI model may be removed or simplified (e.g., pruning less important neurons in neural networks) [21-23] to lower computational overhead with minimal impact on overall system reliability.

**Carbon-Aware Scheduling Strategies:** To reduce environmental impact without significantly impacting overall reliability, energy-heavy workloads could be scheduled to run when renewable, clean energy (e.g., solar, wind) is abundant or available [24]. By factoring in carbon intensity in task scheduling, the AI system may also operate more sustainably.

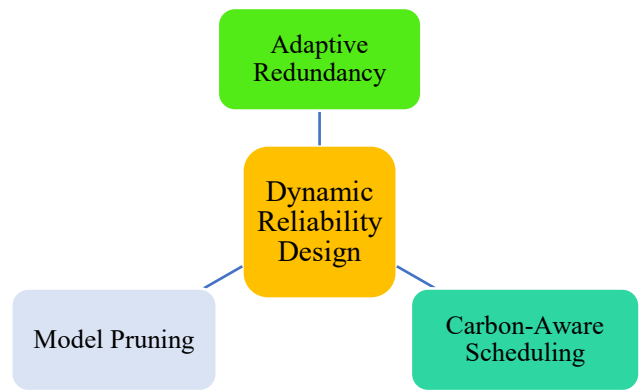

Fig. 6 Key Research Components of Resource-Aware Reliability Design

Each component explained above represents a promising research direction, contributing to energy-aware reliability designs that balance reliability and efficiency and making agentic AIs more feasible for energy-constrained environments and more sustainable in general.

## 7 Reliability Testing and Evaluation

While traditional software systems benefit from mature and standardized reliability testing methodologies (e.g., ISO/IEC/IEEE 29119) [25, 26], these techniques fall short when applied to agentic AI systems. The inherent non-determinism and dynamic action space along with other complex aspects of agentic AI systems (summarized in Table 1) demand new testing and evaluation paradigms.

To establish a robust testing and evaluation paradigm for agentic AI systems, several key research directions can be pursued, including

- Developing a behavioral testing framework where conflicting goals, noisy or adversarial inputs, or unexpected tool failures can be injected to observe and evaluate the system's resilience and adaptability under uncertainty.

- Developing logging and tracing tools that record not only memory reads/writes and tool calls/return values but also reasoning chains and plan/action updates.

- Designing and automating a multi-dimensional reliability evaluation framework that assess failure resilience, explanation quality, consistency, plan efficiency, and safety compliance.
- Developing and applying formal methods (e.g., model checking) to verify agent plans and behaviors for AI systems in high-assurance application domains like aviation and healthcare.

**Table 1.** Differences between Traditional and Agentic AI systems

| Criterion | Traditional Systems | Agentic AI |
|---|---|---|
| Determinism | High (same output for the same input) | Low (stochastic behaviors; output may vary on the same prompt) |
| Action space | Explicitly defined and bounded | Large and dynamic; unexpected emergent behaviors |
| Reproducibility | High | Low (different plans or paths across runs) |
| Specification | Clearly defined function requirements | May be ambiguous, underspecified, or emergent |
| Failure modes | Often discrete and classifiable | Often context-dependent and can be cascading |
| Evaluation metrics | Objective criteria (accuracy, efficiency) | Ambiguous, may involve human-in-the-loop and multi-dimensional evaluation (task completion, safety compliance, user satisfaction) |
| Logging & tracing | Easy | Often difficult due to opaque reasoning and interactions with external tools |

## Acknowledgements

This chapter partially draws upon insights presented in Chapters 1 and 10. We would like to acknowledge the valuable work of these chapter authors whose findings and perspectives have contributed to challenges and future directions in the field offered in this chapter. This work was partially supported by the U.S. National Science Foundation under Grant No. 2302094.

# References

1. Turing, A. M. (1950). Computing Machinery and Intelligence. Mind 49: 433-460
2. Delipetrev, Blagoj and Tsinaraki, Chrysi and Kostic, Uros (2020) Historical Evolution of Artificial Intelligence. Technical Report. Publications Office of the European Union.
3. Joshi, S. (2025) Model Risk Management in the Era of Generative AI: Challenges, Opportunities, and Future Directions," Int. J. Sci. Res. Publ. 15(5): 299–309, doi: 10.29322/ijsrp.15.05.2025.p16133
4. Liao, W., Salinas, S., Li, M., Li, P., and Loparo, K. (2017). Cascading failure attacks in the power system: A stochastic game perspective. IEEE Internet Things J. 4(6): 2247–2259.
5. Hu, P., Mei, T., Fan, W. (2017) Cascading failure forecast of complex power grids: A review. Scientia Sinica Technologica 47(4): 355–363.
6. Liu, D. and Tse, C.K. (2019) Cascading Failure of Cyber-Coupled Power Systems Considering Interactions Between Attack and Defense. IEEE Transactions on Circuits and Systems I: Regular Papers 66(11): 4323-4336, doi: 10.1109/TCSI.2019.2922371
7. Zheng, D., Fu, X., Liu, X., Xing, L., Peng, R. (2025) Modeling and Analysis of Cascading Failures in Industrial Internet of Things Considering Sensing-Control Flow and Service Community. IEEE Transactions on Reliability 74(2): 2723-2737.
8. Fu, X., Zheng, D., Liu, X., Xing, L., Peng, R. (2025) Systematic review and future perspectives on cascading failures in Internet of Things: Modeling and optimization. Reliability Engineering & System Safety 254, Part A, 110582.
9. Xing, L. (2024) Reliability and Resilience in the Internet of Things. Elsevier.
10. Xing, L. (2021) Cascading Failures in Internet of Things: Review and Perspectives on Reliability and Resilience. IEEE Internet of Things Journal 8(1): 44-64.
11. Allam, H., AlOmar, B., Dempere, J. (2025) Agentic AI for IT and Beyond: A Qualitative Analysis of Capabilities, Challenges, and Governance. Artif. Intell. Bus. Rev. 2025, 1. Available online: https://theaibr.com/index.php/aibr/article/view/3
12. Bandi, A., Kongari, B., Naguru, R., Pasnoor, S., & Vilipala, S. V. (2025) The Rise of Agentic AI: A Review of Definitions, Frameworks, Architectures, Applications, Evaluation Metrics, and Challenges. Future Internet 17(9), 404. https://doi.org/10.3390/fi17090404
13. Spirtes, P., Glymour, C., Scheines, R. (2001) Causation, Prediction, and Search. The MIT Press. URL: https://direct.mit.edu/books/book/2057/causation-prediction-and-search. doi: 10.7551/mitpress/1754.001.0001 .
14. Pearl, J. (2011) Causality: Models, reasoning, and inference, second edition, doi:10. 1017/CBO9780511803161.
15. Fournier, F., Limonad, L., David, Y. (2025) Agentic AI Process Observability: Discovering Behavioral Variability, https://doi.org/10.48550/arXiv.2505.20127
16. Ouyang, S., Zhang, JM., Harman, M., Wang, M. (2025) An empirical study of the non-determinism of chatgpt in code generation. ACM Trans. Softw. Eng. Methodol. 34, https://doi.org/10.1145/3697010. doi:10.1145/3697010.
17. Atil, B., Aykent, S., Chittams, A., Fu, L., Passonneau, R. J., Radcliffe, E., Rajagopal, GR., Sloan, A., Tudrej, T., Ture, F., Wu, Z., Xu, L., Baldwin, B. (2025) Non-determinism of "deterministic" LLM settings, https://arxiv.org/abs/2408.04667. arXiv:2408.04667.
18. Garg, V. (2025). Designing the Mind: How Agentic Frameworks Are Shaping the Future of AI Behavior. Journal of Computer Science and Technology Studies 7(5): 182-193. https://doi.org/10.32996/jcsts.2025.7.5.24
19. Murugesan, S. (2025) The Rise of Agentic AI: Implications, Concerns, and the Path Forward. IEEE Intelligent Systems 40(2): 8-14, doi: 10.1109/MIS.2025.3544940.
20. Johnson, B. W. (1989) Design and Analysis of Fault Tolerant Digital Systems. Addison-Wesley.
21. Mohanty, L., Kumar, A., Mehta, V. et al. (2025) Pruning techniques for artificial intelligence networks: a deeper look at their engineering design and bias: the first review of its kind. Multimed Tools Appl 84: 9591–9665, https://doi.org/10.1007/s11042-024-19192-x

22. Jia, Y., Liu, B., Zhang, X., Dai, F, Khan, A, Qi, L, Dou, W. (2025) Model Pruning-enabled Federated Split Learning for Resource-constrained Devices in Artificial Intelligence Empowered Edge Computing Environment. ACM Transactions on Sensor Networks, in press, https://doi.org/10.1145/3687478
23. Yeom, S., Seegerer, P., Lapuschkin, S., Binder, A., Wiedemann, S., Müller, K., Samek, W. (2021) Pruning by explaining: A novel criterion for deep neural network pruning. Pattern Recognition 115, 107899, https://doi.org/10.1016/j.patcog.2021.107899.
24. Hewage, TB, Ilager, S., Rodriguez MA, and Buyya, R. (2025) A Framework for Carbon-Aware Real-Time Workload Management in Clouds Using Renewables-Driven Cores. IEEE Transactions on Computers 74(8): 2757-2771.
25. Pham, H. (1995) Software Reliability and Testing. IEEE Computer Society Press, Washington, DC, United States.
26. Homès, B. (2024) Fundamentals of Software Testing, 2nd edition, Wiley.